\pdfoutput=1

\documentclass[11pt]{article}

\usepackage[preprint]{acl}

\usepackage{times}
\usepackage{latexsym}
\usepackage{enumitem}

\usepackage[T1]{fontenc}

\usepackage[utf8]{inputenc}

\usepackage{microtype}
\usepackage{lineno}
\usepackage{pifont}
\usepackage{multirow}
\usepackage{amssymb}
\usepackage{array}
\usepackage{amsmath}
\usepackage{graphicx}
\usepackage{adjustbox}
\usepackage{booktabs} 
\DeclareMathOperator*{\argmax}{argmax}
\usepackage{inconsolata}

\usepackage{graphicx}

\title{Diagram-Driven Course Questions Generation}

\author{Xinyu Zhang$^{1,2}$,\; Lingling Zhang$^{1,2}$\thanks{Corresponding author},\; Yanrui Wu$^{1,2}$,\;Muye Huang$^{1,2}$,\;Wenjun Wu$^{1,2}$, \\
\textbf{Bo Li,}$^{1,2}$\; \textbf{Shaowei Wang,}$^{1,3}$\; \textbf{Basura Fernando,}$^{4,5}$\;
\textbf{Jun Liu}$^{1,3}$
\\
{$^{1}$School of Computer Science and Technology, Xi’an Jiaotong University}\; \\
{$^{2}$Ministry of Education Key Laboratory of Intelligent Networks and Network Security, China} \; \\
{$^{3}$Shaanxi Province Key Laboratory of Big Data Knowledge Engineering, China} \; \\
{$^{4}$IHPC, Agency for Science, Technology and Research, Singapore} \; \\
{$^{5}$College of Computing and Data Science, Nanyang Technological University, Singapore} \; \\
\texttt{{zhang1393869716}@stu.xjtu.edu.cn, \{zhanglling,liukeen\}@xjtu.edu.cn}
}
\begin{document}
\maketitle
\begin{abstract}
Visual Question Generation (VQG) research focuses predominantly on natural images while neglecting the diagram, which is a critical component in educational materials.
To meet the needs of pedagogical assessment, we propose the Diagram-Driven Course Questions Generation (DDCQG) task and construct DiagramQG, a comprehensive dataset with 15,720 diagrams and 25,798 questions across 37 subjects and 371 courses.
Our approach employs \textit{course} and \textit{input} text constraints to generate course-relevant questions about specific diagram elements.
We reveal three challenges of DDCQG: domain-specific knowledge requirements across courses, long-tail distribution in course coverage, and high information density in diagrams.
To address these, we propose the Hierarchical Knowledge Integration framework (HKI-DDCQG), which utilizes trainable CLIP for identifying relevant diagram patches, leverages frozen vision-language models for knowledge extraction, and generates questions with trainable T5.
Experiments demonstrate that HKI-DDCQG outperforms existing models on DiagramQG while maintaining strong generalizability across natural image datasets, establishing a strong baseline for DDCQG.
\end{abstract}

\section{Introduction}
\label{sec:intro}
Visual Question Generation (VQG) represents a pivotal and promising research domain with significant educational applications \cite{xie2025explicitly, luo-etal-2024-chain}.
While VQG focuses on automatically generating questions from visual inputs, current research predominantly addresses natural images while neglecting diagrams \cite{wang2024cvpr, zhang-etal-2025-physreason, 10812784}, a fundamental component of educational materials. 
This critical gap impedes the advancement of deep learning technologies in educational contexts.
\par
The diagram plays a crucial role in pedagogical assessment by facilitating structured information presentation and evaluating students' comprehension of knowledge \cite{cook2006visual}. 
Therefore, we propose the \textbf{Diagram-Driven Course Questions Generation} (DDCQG) task, which aims to encourage models to generate questions based on diagrams for specific courses.
These questions are essential for evaluating students' abilities to explain, analyze, and apply course knowledge based on the diagrams \cite{lambertus2008students}.
\begin{figure}[t]
\centering
\includegraphics[width=0.42\textwidth]{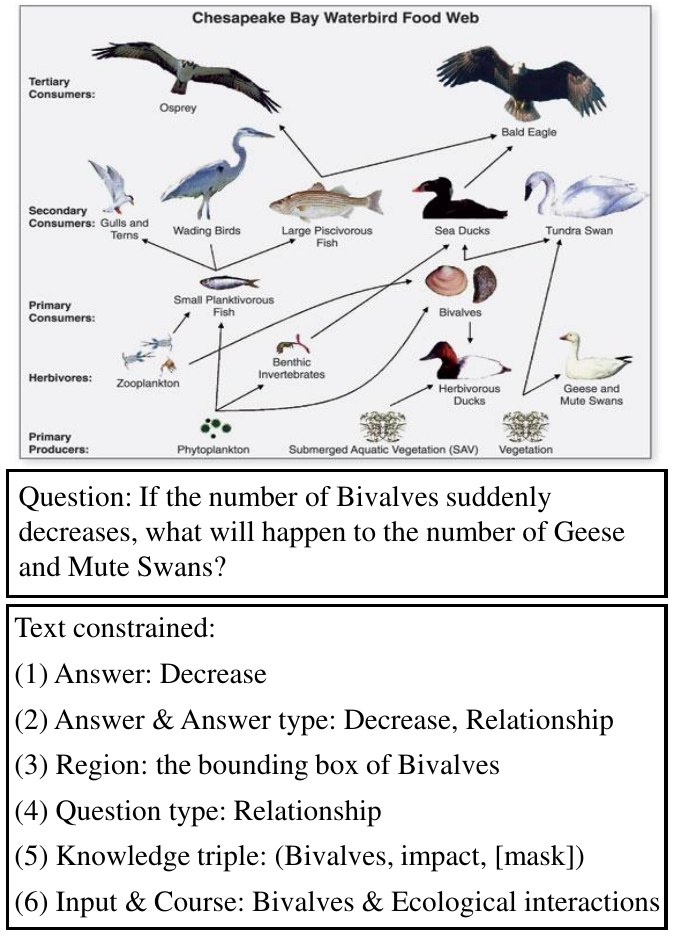}
\vspace{-5pt}
\caption{Comparison of different text constraints for an example in DiagramQG.}
\label{fig:0}
\vspace{-15pt}
\end{figure}
\par
Furthermore, current VQG research extensively employs various text constraints—including answers \cite{xie2025explicitly, mi2024convqg}, answer types \cite{krishna2019information}, image regions \cite{uehara2018visual}, question types \cite{fan2018question}, and knowledge triples \cite{uehara2023k}. 
However, these constraints face significant limitations when applied to DDCQG task.
Research demonstrates that existing approaches frequently produce context-independent questions \cite{liu-etal-2024-knowledge}, fail to align with intended assessment objectives \cite{uehara2023k}, or merely generate superficial expansions lacking pedagogical depth \cite{mi2024convqg}, highlighting the need of specific text constraint for DDCQG task.
\par
To address these issues, we present DiagramQG, a comprehensive dataset comprising 15,720 diagrams and 25,798 questions spanning 6 disciplines, 37 subjects, and 371 courses. 
We propose a novel text constraint with \textit{course} and \textit{input}, as shown in Figure \ref{fig:0}, where \textit{course} constraint ensure question relevance to specific educational contexts, and \textit{input} constraint guide question generation around targeted diagram elements.
Through analysis of DiagramQG, we identify three fundamental challenges of DDCQG: \textbf{domain-specific knowledge requirement}, with models needing to possess specialized knowledge across various disciplines unlike existing VQG datasets that rely primarily on general common sense; \textbf{long-tail distribution phenomenon}, where course coverage ranges from abundant to limited, challenging models to generalize across both well-represented and underrepresented courses; and \textbf{high object information density}, as diagrams contain concentrated visual information that complicates content interpretation and risks overlooking critical details.
\par
To address these challenges, we propose the Hierarchical Knowledge Integration framework for DDCQG task (HKI-DDCQG) as a strong baseline.
This framework employs a trainable CLIP to identify relevant multi-scale diagram patches, utilizes vision-language models like BLIP \cite{li2022blip} and Qwen2.5-VL \cite{bai2025qwen2} for knowledge extraction, and implements T5 for filtering extracted knowledge to generate the question based on text constraints.
The framework then integrates these filtered insights with text constraints and multi-scale diagram patches for question generation.
Notably, this frame freezes the VLM's parameters and trains only the remaining parts, thus improving scalability and cost-efficiency.
\par
We evaluate HKI-DDCQG against existing VQG and vision-language models on our DiagramQG dataset and validate its generalizability through experiments on four natural image VQG datasets.
Our primary contributions include the following:
\begin{itemize}[leftmargin=*,itemsep=0pt,parsep=0pt,topsep=0pt]
    \item We construct the DiagramQG dataset, incorporating course and input constraints to guide the generation of Diagram-Driven Course Questions that align with educational requirements.
    \item We propose the HKI-DDCQG, which leverages frozen-parameter vision-language models to enable cost-effective diagram-driven course question generation. The framework demonstrates excellent performance across both DiagramQG and various natural image VQG datasets.
    \item We conduct comprehensive evaluations of mainstream VQG models, diverse vision-language models, and our HKI-DDCQG on the novel DiagramQG dataset, establishing new benchmarks for diagram-based question generation.
\end{itemize}
\begin{figure*}[t]
    \centering
    \includegraphics[width=0.97\textwidth]{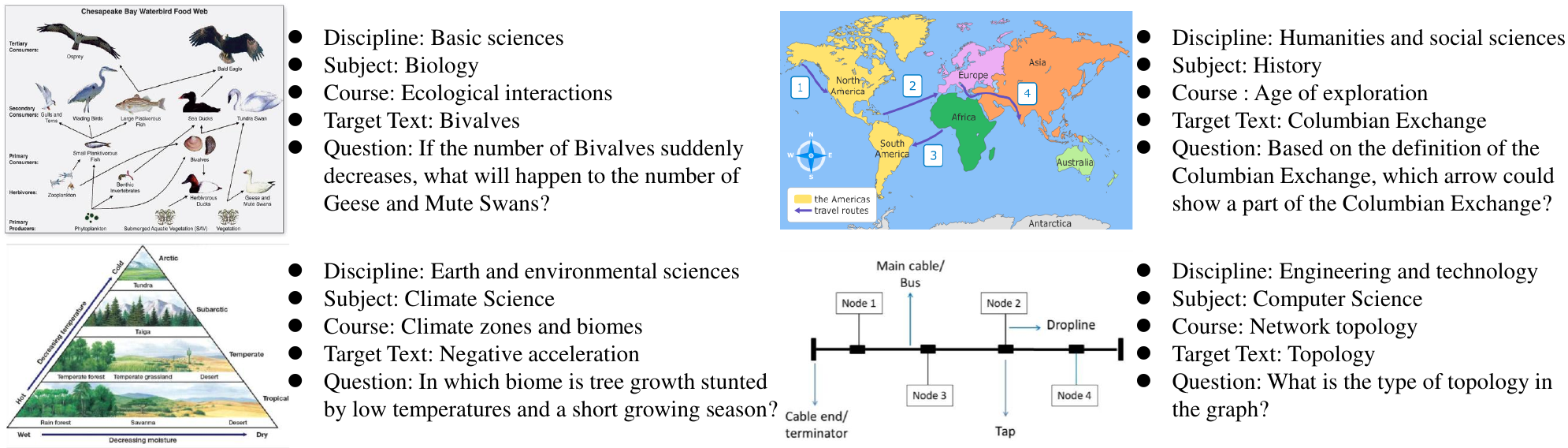}
    \vspace{-5pt}
    \caption{Four different examples of different subjects in DiagramQG dataset.}
    \label{fig:1}
    \vspace{-10pt}
\end{figure*}
\begin{figure*}[t]
    \centering
    \includegraphics[width=0.97\textwidth]{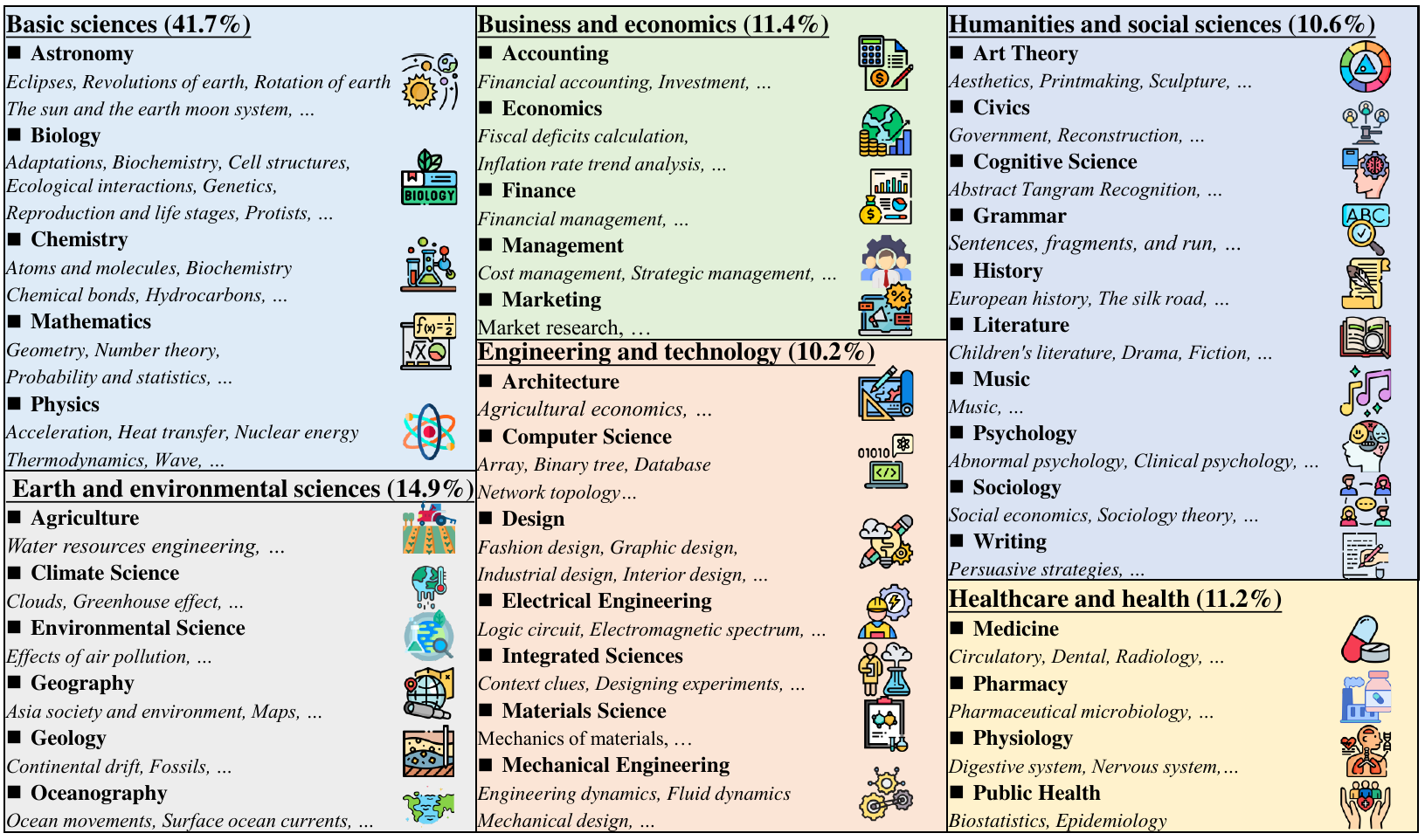}
     \vspace{-5pt}
    \caption{Domain diversity in DiagramQG where each color corresponds to one discipline. }
    \label{fig:2}
    \vspace{-10pt}
\end{figure*}
\section{Related Work}
\label{sec:related}
\subsection{Visual Question Generation}
VQG has evolved from rule-based approaches to sophisticated neural architectures.
While unconstrained approaches often generate generic questions lacking image specificity \cite{bi2022inferential}, constrained methods have shown success through various strategies, including answer guidance \cite{xu2020radial, liu-etal-2024-look}, knowledge enhancement \cite{xie2022knowledge, chen2023deconfounded}, cross-topic models \cite{liu2024knowledge}, and contrastive learning \cite{mi2024convqg}.
There are also some studies \cite{xie2025explicitly} on the generation of visual problems with controllable difficulty.
However, challenges persist in balancing question diversity with effective visual information utilization, necessitating innovative constraint mechanisms.

\subsection{Diagram Question Answering}
Diagram Question Answering (DQA) often demands enhanced reasoning capabilities and domain knowledge. 
Prior research has focused on improving diagram comprehension through pre-training methods \cite{gomez2020isaaq, ma2022weakly, XU2023109588} and utilizing Large Language Models via advanced prompting techniques \cite{lu2022learn, zhang2023multimodal, yao2024tree, wang2024cvpr, huang2025vprochart, huang2025evochart}. 
The effective integration of visual diagrammatic information with background knowledge remains absolutely crucial for improving DQA performance.

\section{DiagramQG Dataset}
\subsection{Data Collection}
The data collection process consisted of three distinct phases. 
First, we gathered diagrams and related questions from existing diagram-related datasets \cite{kembhavi2017you, wang2022computer, zhang2024alignment, lu2022learn, yue2024mmmu, chen2024m3cot}, supplemented by diagrams available for academic use from platforms such as Hugging Face and Roboflow. 
This initial phase yielded a substantial raw dataset containing over 25,000 diagrams and 60,000 questions.
\par
Subsequently, we organized the collected diagrams and questions into six primary disciplines and further categorized them into 37 specific subjects. This structuring process involved mapping questions to their corresponding courses, resulting in the identification of 371 distinct courses.
\par
Finally, highly experienced subject-specific annotators with relevant knowledge backgrounds annotated input text constraints for each diagram-question pair within their specialized domains.
A separate group of subject annotators evaluated all samples based on their educational utility using a 100-point scale.
Samples scoring below the threshold of 70 are removed, and only the highest-scoring set of text constraints is retained for each diagram-question pair.
The resulting DiagramQG dataset contains 25,798 samples associated with 15,720 unique diagrams. Examples from four subjects in the DiagramQG dataset are shown in Figure \ref{fig:1}.
\par
\subsection{Data Analyse}
\subsubsection{Domain \& Question diversity.}
\begin{table*}
\centering
\caption{Comparing characteristics of other datasets and DiagramQG, where Q. and I. mean question and image.}
\begin{adjustbox}{width=\textwidth}
\begin{tabular}{ccccccc}
\hline
 & Num. of Q. & Num. of I. & Object on I.  & Images & Text Constraint & Subjects\\
\hline
VQAv2.0 & 1.1M & 20k  & 3.5 & natural & answer & N/A \\
FVQA & 5,826 & 2k  & 2.9 & natural & answer & N/A \\
VQG-COCO & 25,000 & 5k & 3.3 & natural & image caption & N/A \\
K-VQG & 16,098 & 13K & 2.7 & natural & knowledge triple &  N/A \\
\hline
TQA & 26,260 & 3,455 & 7.2 & diagram & - &  10\\
CSDQA & 3,494 & 1,294 & 5.2 & diagram & - &  1\\
ADE & 3.945 & 3.945 & 7.4 & diagram & - &  8\\
ScienceQA & 21,208 & 10,332 & 4.3 & diagram \& natural & - &  12 \\
MMMU & 11,550 & 11,264 & 4.5 & diagram \& natural & - & 30 \\
M3CoT & 11,459 & 11,293 & 5.4 & diagram \& natural & - &  17 \\
\hline
DiagramQG & 25,798 & 15,720 & 10.5 & diagram & input, course & 37 \\
\hline
\end{tabular}
\end{adjustbox}
\label{tab:1}
\vspace{-10pt}
\end{table*}
As illustrated in Figure \ref{fig:2}, the DiagramQG dataset encompasses 6 disciplines, 37 subjects, and 371 courses across multiple academic domains, with an average of 17.45 words per question.
The dataset follows a hierarchical structure, with samples first classified by discipline, then divided into specific subjects (e.g., Biology), and ultimately categorized into courses (e.g., Ecological interactions).
For further research, we split DiagramQG into train, val, and test sets with a ratio of 70:5:25. Considering that some courses contain fewer than 5 samples, we prioritized allocating these courses to the test set to ensure a comprehensive evaluation.
\par
\begin{figure}[t]
    \centering
    \includegraphics[width=0.48\textwidth]{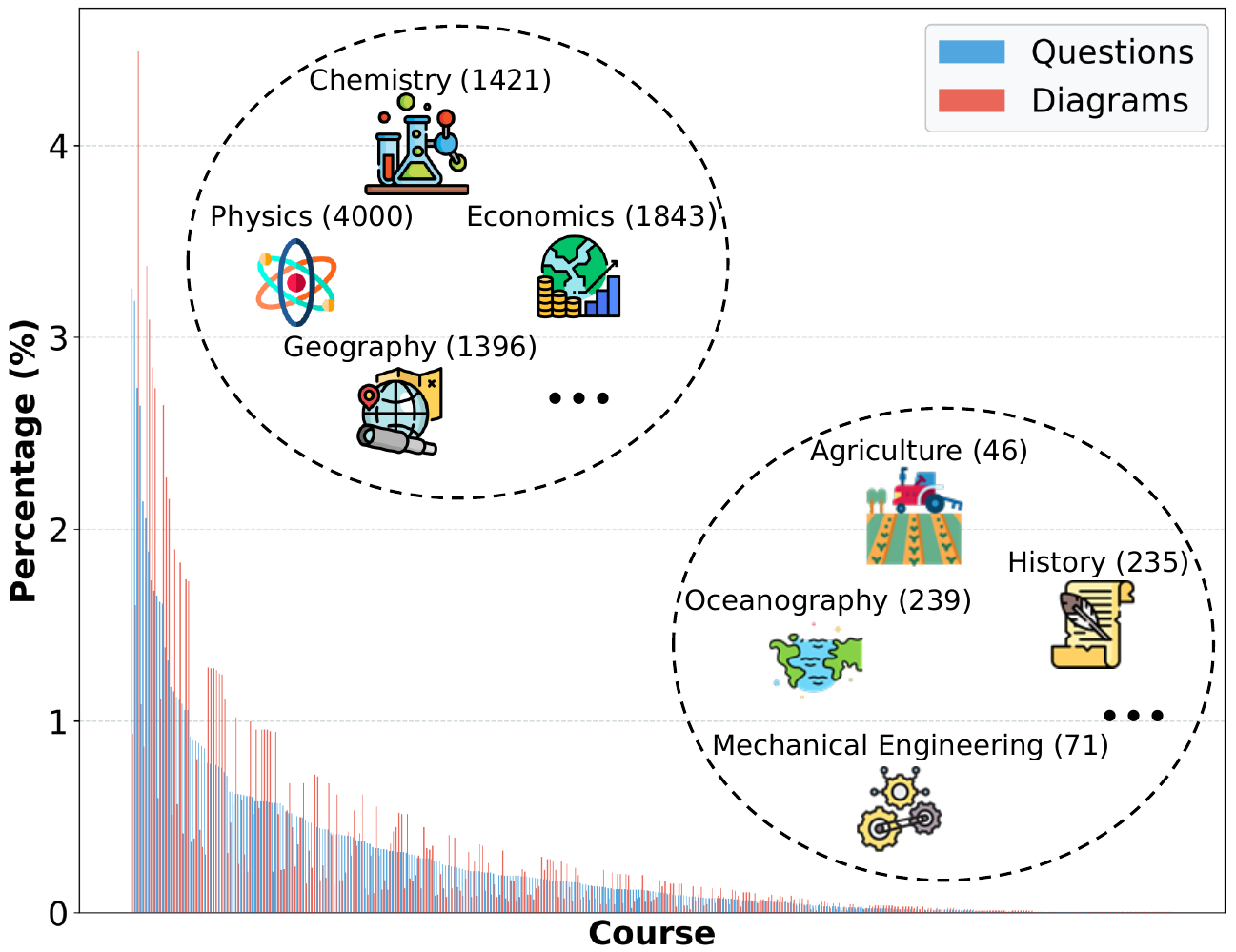}
    \vspace{-10pt}    
    \caption{Distribution of diagrams, questions ratios across different courses in DiagramQG.}
    \label{fig:5}
    \vspace{-15pt}
\end{figure}

As illustrated in Figure \ref{fig:5}, the ratio of questions and diagrams for each course exhibits a pronounced long-tail distribution.
This asymmetric distribution indicates that the DiagramQG dataset covers a wide range of courses, some courses have notably limited samples.
However, this is a common characteristic in educational scenarios that accurately reflects typical resource allocation patterns in real-world educational environments.
\subsubsection{Comparisons to Other Datasets}
Table \ref{tab:1} presents a comprehensive comparison between DiagramQG and existing VQG-related and diagram-related datasets, highlighting DiagramQG's distinctive characteristics. 
DiagramQG encompasses 25,798 questions and 15,720 unique diagrams, substantially surpassing the scale of existing multidisciplinary datasets (such as M3CoT), including those that incorporate both natural images and diagrams.
Unlike conventional VQG datasets that primarily focus on image caption and common-sense reasoning, DiagramQG is specifically engineered for educational applications.
\par
DiagramQG demonstrates three significant advantages. 
First, it exhibits unprecedented educational breadth, spanning 37 distinct academic courses, which exceeds the domain coverage of existing datasets. 
Second, the dataset introduces a novel constraint that integrates both input phrases and course-specific contextual information, transcending traditional constraint formats. 
Third, each diagram contains an average of 10.5 objects, representing significantly higher complexity compared to existing datasets in the field.
This unique combination of comprehensive course coverage, sophisticated constraint mechanisms, and enhanced diagram complexity establishes DiagramQG as the first extensive dataset specifically designed for diagram question generation across diverse educational domains. 
The dataset facilitates the development of more robust and versatile question-generation systems for educational applications.

\subsubsection{Challenges in DiagramQG Dataset}
Our comparative analysis reveals three distinctive challenges in DiagramQG that set it apart from existing visual question generation datasets:
\begin{itemize}[leftmargin=*,itemsep=0pt,parsep=0pt,topsep=0pt]
\item \textbf{Domain-specific knowledge requirement:} Unlike other existing VQG datasets that focus on general common sense, DiagramQG dataset consistently requires models to possess and apply different courses across various subjects to generate meaningful and practical questions.
\item \textbf{Long-tail distribution phenomenon:} The inherent complex long-tail distribution in DiagramQG dataset, where course samples range from abundant to limited, challenges the generalization and performance of models across both sample-rich and sample-limited courses.
\item \textbf{High object information density:} The significantly high density of object information in the diagrams complicates content interpretation and risks overlooking critical details, demanding models to possess capabilities in capturing and processing complex visual information.
\end{itemize}
\section{Baseline}
\subsection{Problem Definition}
The Diagram-Driven Course Questions Generation (DDCQG) task aims to generate a pedagogically appropriate question $q$ given a diagram $d$, an input text $t$, and a course text $c$ specifically.
The generated questions are used to effectively assess students' understanding of the specified course $c$.
This task can be formulated as a conditional generation problem, represented as $p(q|d,t,c)$, where a multimodal model coherently maps visual and textual information into a joint embedding space before decoding questions that satisfy both the text constraint and the course assessment requirement.
\begin{figure*}[t]
    \centering
    \includegraphics[width=0.96\textwidth]{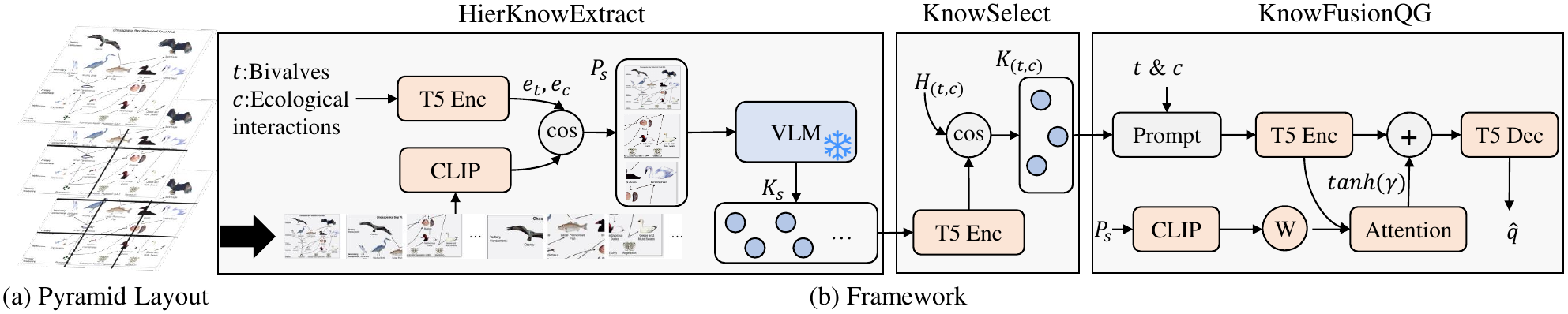}
    \caption{(a) Process the diagram into a pyramid structure of patches.
    (b) Our DiagramQA baseline, HKI-DDCQG, consists of three distinct stages: HierKnowExtract, KnowSelect, and KnowFusionQG.
    In this framework, orange modules indicate trainable parameters, blue modules represent fixed parameters, and gray modules denote components without learnable parameters.}
    \label{fig:6}
    \vspace{-10pt}
\end{figure*}
\subsection{Architecture}
We propose the Hierarchical Knowledge Integration (HKI-DDCQG) framework as a baseline for the DDCQG task.
This framework is designed to be compatible with any vision-language foundation model and implements a three-stage pipeline for question generation: \textbf{HierKnowExtract}, \textbf{KnowSelect}, and \textbf{KnowFusionQG}, as shown in Figure \ref{fig:6}.
The \textbf{HierKnowExtract} stage extracts knowledge $K_s$ from multi-scale diagram patches $P_s$ using vision-language models with frozen parameters.
The \textbf{KnowSelect} stage selects the $m$ most relevant knowledge sentences $K_{(t,c)}$ based on text $t$ and course $c$.
The \textbf{KnowFusionQG} stage integrates $t$, $c$, $P_s$, and $K_{(t,c)}$ to generate the final question $\hat{q}$.
\subsubsection{HierKnowExtract}
The \textbf{HierKnowExtract} stage obtains diagram patches $P_s$ of different scales related to the input \& course and uses vision-language models to extract the knowledge $K_s$ contained in all patches. 
This process begins with a hierarchical decomposition of diagram $d$ into ordered patches $P_d$ across an $n$-layered pyramid structure, followed by visual encoding using the CLIP Image Encoder \cite{radford2021learning}  to get $F^l$, as formulated in Equation \ref{eq:0}.
\begin{equation}
\left\{
\begin{aligned}
& F_d=\{F^l\}_{l=1}^n \\ 
& P_d = \{p^{l}_{i,j} \mid l \in [1,n], i,j \in [1,l]\} \\
& F^l = \{f^l_{i,j} = \text{CLIP}(p^l_{i,j}) \mid p^l_{i,j} \in P_d\} \\
& p^{l}_{i,j} = d\left[\frac{i-1}{l}H:\frac{i}{l}H, \frac{j-1}{l}W:\frac{j}{l}W\right] \\
\end{aligned}
\right.
\label{eq:0}
\end{equation}
where $H$ and $W$ denote the height and width of the input diagram, respectively.
\par
For textual components, the T5 Encoder \cite{raffel2020exploring} is employed to process both the input text $t$ and course text $c$ independently. 
We introduce a learnable linear transformation $W_h$ to ensure seamless compatibility between the CLIP Image Encoder \cite{radford2021learning} and T5 Encoder \cite{raffel2020exploring} feature spaces. 
This effectively facilitates unified similarity computation and subsequent operations in the \textbf{KnowFusionQG} phase.
The process ultimately culminates in the selection of the most semantically relevant patch from each hierarchical layer to form the patch set $P_s$, as shown in Equation \ref{eq:1}.
\begin{equation}
\left\{
\begin{aligned}
& e_t = \text{T5}_{\text{enc}}(t), \quad e_c = \text{T5}_{\text{enc}}(c) \\
& s^l_{i,j} = \text{sim}(W_h f^l_{i,j}, e_t) + \text{sim}(W_h f^l_{i,j}, e_c) \\
& P_s = \{p^l_{i^*,j^*} \mid (i^*,j^*) = \argmax_{i,j} s^l_{i,j}, l \in [1,n]\}
\end{aligned}
\right.
\label{eq:1}
\end{equation}
The selected patches $P_s$, input text $t$, and course text $c$ are carefully fed into a large-scale vision-language model (VLM) like Qwen2.5-VL \cite{bai2025qwen2} or BLIP \cite{li2022blip} to obtain diverse knowledge which is related with patch, input and course.
The resulting knowledge set $K_s$ is systematically generated according to Equation \ref{eq:2}.
\begin{equation}
\left\{
\begin{aligned}
& K_s = \{k^l \mid l \in [1,n]\} \\
& k^l = \text{VLM}(p^l_{i^*,j^*}, t, c), \quad \forall p^l_{i^*,j^*} \in P_s
\end{aligned}
\right.
\label{eq:2}
\end{equation}
where $k^l$ is the text paragraph retrieved by the VLM, containing several knowledge sentences.
To optimize the learning process, the CLIP Image Encoder \cite{radford2021learning}, T5 Encoder \cite{raffel2020exploring}, and the linear transformation $W_h$ employed in this phase share parameters with their counterparts in the third phase, where gradient updates are effectively propagated.

\subsubsection{KnowSelect}
The \textbf{KnowSelect} stage selects the $m$ most relevant knowledge sentences $K_{(t,c)}$ from the extensive knowledge set $K_s$ based on input text $t$ and course text $c$. 
The process begins with encoding the knowledge set $K_s$ and the $t,c$ text constraints using the T5 Encoder \cite{raffel2020exploring}, as formulated in the following Equation \ref{eq:3}.
\begin{equation}
\left\{
\begin{aligned}
H_K &= \text{T5}_{\text{enc}}(K_s) \\
H_{t,c} &= \text{T5}_{\text{enc}}(\text{Prompt}(t,c))
\end{aligned}
\right.
\label{eq:3}
\end{equation}
where $\text{Prompt}(t,c)$ combines $t$ and $c$ into a prompt following the template: \textit{Given the input text {$t$}, identify key knowledge related to the course {$c$}}.
\par
For the semantic relevance between the knowledge set ($K_{(t,c)}$) and text constraints ($t$ and $c$), we employ a scaled dot-product attention mechanism. 
This computes attention scores between the knowledge tokens and the text constraint, followed by knowledge selection, as formulated in Equation \ref{eq:4}:
\begin{equation}
\left\{
\begin{aligned}
& A = \text{softmax}\left(\frac{H_K H_{t,c}^T}{\sqrt{d_k}}\right) \\
& K_{(t,c)} = \text{top-m}(A H_K)
\end{aligned}
\right.
\label{eq:4}
\end{equation}
where $d_k$ denotes the dimensionality of the hidden representations.
The $\text{top-m()}$ operator selects the $m$ most semantically relevant knowledge sentences based on attention scores, creating a knowledge set $K_{(t,c)}$ that best matches the text constraints.

\subsubsection{KnowFusionQG}
The \textbf{KnowFusionQG} stage integrates the selected diagram patches $P_s$, input text $t$, course text $c$, and knowledge set $K_{(t,c)}$ to generate the diagram-driven course question through a multimodal fusion mechanism. 
This process begins with encoding the visual and textual inputs into language and vision representations, as shown in Equation \ref{eq:7}.
\begin{equation}
\left\{
\begin{aligned}
& H_{t} = \text{T5}_{\text{enc}}(\text{Prompt}[t;c;K_{(t,c)}]) \\
& H_{v} = W_h \cdot \text{CLIP}(P_s)
\end{aligned}
\right.
\label{eq:7}
\end{equation}
where $W_h$ is also used in \textbf{HierKnowExtract} phase. 
This $\text{Prompt}[t;c;K_{(t,c) }]$ synthesizes $t$, $c$ and $K_{(t,c)}$ into a coherent prompt following the template: \textit{Generate the question including input {$t$} to assess course {$c$} with the knowledge {$K_{(t,c)}$}}.
To capture the intricate relationships between textual and visual representations, a cross-modal attention mechanism followed by a gated fusion network is employed, as formulated in Equation \ref{eq:8}.
\begin{equation}
\left\{
\begin{aligned}
& H^{attn}_{v} = \text{Softmax}(\frac{H_{t}H_{v}^T}{\sqrt{d_k}})H_{v} \\
& \lambda = W_t H_{t} + W_v H^{attn}_{v} \\
& H_{fuse} =  H_{t} + \tanh(\lambda) \cdot H^{attn}_{v}
\end{aligned}
\right.
\label{eq:8}
\end{equation}
Finally, the fused output $H_{fuse}$ is fed into the T5 decoder \cite{raffel2020exploring} to predict the input question $\hat{q}$. 
This integration of visual and text information through cross-modal attention and gated fusion enables the model to generate questions that are contextually relevant and conceptually focused.

\section{Experiments}
\subsection{Implementation Details}
The experimental framework uses T5-Base and T5-Large architectures along with the CLIP (ViT-B/32) model.
Optimization is performed with the AdamW optimizer, applying a learning rate of 1e-5 for CLIP and 5e-5 for others, running through 10 epochs of fine-tuning. 
Key parameters include a maximum input sequence length of 256, an output sequence length of 64, and a batch size of 32, with experiments conducted on one NVIDIA A800 80G GPU.
The DiagramQG is split into training, validation, and testing sets with a ratio of 70:5:25, ensuring no overlap of diagrams and questions across these sets.
Additionally, each course is represented in both validation and testing sets, despite some courses having relatively few samples.
\begin{table*}
\centering
\caption{Results on DiagramQG, where GPT-4o (T) and Claude-3.5-Sonnet (T) are meant to generate questions using only text, where, Q-3B, Q-7B, T5-B, T5-L mean, Qwen2.5-VL-3B, Qwen2.5-VL-7B, T5-Base and T5-Large.}
\begin{adjustbox}{width=\textwidth}
\begin{tabular}{lccccccccc}
\hline
Method & BLEU-1 & BLEU-2 & BLEU-3 & BLEU-4 & Bert-Score & METEOR & CIDEr & ROUGE-L & Fleur\\
\hline
\textbf{\textit{In-Context Learning}} \\
InternVL2.5 (26B) \cite{wang2025internvideo2} & 32.52 & 19.78 & 13.14 & 9.52 & 92.37 & 18.51 & 0.42 & 28.10 & 60.31\\
InternVL2.5 (38B) \cite{wang2025internvideo2} & 33.34 & 21.15 & 14.68 & 10.68 & 93.76 & 20.13 & 0.44 & 29.66 & 62.13\\
MiniCPM-V2(8B) \cite{yao2024minicpm} & 28.37 & 15.34 & 9.66 & 6.31 & 88.15 & 14.26 & 0.33 & 25.69 & 55.14\\
DeepSeek-VL(7B) \cite{lu2024deepseek} & 28.27 & 17.13 & 10.68 & 7.27 & 89.88 & 17.85 & 0.35 & 27.17 & 53.75\\
Qwen2.5-VL(3B) \cite{bai2025qwen2} & 31.44 & 18.75 & 11.64 & 7.36 & 88.76 & 15.51 & 0.34 & 27.15 & 58.95\\
Qwen2.5-VL(7B) \cite{bai2025qwen2} & 33.51 & 21.06 & 13.93 & 10.52 & 90.53 & 18.11 & 0.51 & 29.00 & 61.34\\
Qwen2.5-VL(72B) \cite{bai2025qwen2} & 41.91 & 29.45 & 22.16 & 17.56 & 96.86 & 23.21 & 0.80 & 36.57 & 64.15\\
GLM4-V pro & 37.31 & 23.81 & 16.34 & 12.85 & 95.66 & 19.68 & 0.70 & 32.07 & 63.12\\
Claude-3.5-Sonnet (T) & 27.21 & 15.93 & 10.00 & 6.62 & 92.75 & 16.10 & 0.51 & 24.35 & 22.76\\
GPT-4o (T) & 23.18 & 12.78 & 8.04 & 4.74 & 94.82 & 13.60 & 0.41 & 24.22 & 21.25\\
Claude-3.5-Sonnet & 48.98 & 40.66 & 34.68 & 30.07 & 98.21 & 34.62 & 2.05 & 45.23 & 67.69\\
GPT-4o & 51.15 & 43.08 & 37.16 & 32.33 & 98.94 & 35.92 & 2.20 & 49.60 & 65.29\\
\hline
\textbf{\textit{Fine-tuned}} \\
K-VQG \cite{uehara2023k} & 20.54 & 15.31 & 11.98 & 9.58 & 73.66 & 24.76 & 0.41 & 26.29 & 54.25\\
Patch-TRM \cite{lu2021iconqa} & 21.13 & 15.90 & 12.49 & 10.12 & 74.46 & 25.91 & 0.42 & 28.09 & 55.42\\
ConVQG (BLIP) \cite{mi2024convqg} & 28.65 & 21.12 & 16.38 & 13.06 & 74.49 & 26.93 & 0.50 & 28.77 & 55.75\\
KC-VQG (BLIP) \cite{liu-etal-2024-knowledge} & 30.41 & 23.95 & 19.29 & 15.19 & 79.62 & 28.46 & 0.93 & 31.55 & 56.51\\
KC-VQG (Q-3B) \cite{liu-etal-2024-knowledge} & 33.15 & 25.92 & 20.92 & 17.10 & 83.32 & 30.57 & 1.19 & 34.57 & 59.76\\
Qwen2.5-VL-3B (Lora) & 50.75 & 45.37 & 39.36 & 35.52 & 86.04 & 41.72 & 3.05 & 48.21 & 65.78\\
Qwen2.5-VL-7B (Lora) & 53.64 & 48.04 & 41.50 & 37.56 & 88.57 & 44.12 & 3.22 & 50.89 & 68.12 \\
\hline
\textbf{\textit{Ours}} \\
HKI-DDCQG(BLIP, T5-B) & 44.05 & 36.31 & 31.32 & 27.63 & 84.02 & 37.55 & 2.62 & 42.65 & 58.87 \\
HKI-DDCQG(Q-3B, T5-B) & 54.55 & 48.04 & 43.42 & 40.90 & 88.86 & 47.61 & 3.74 & 54.17 & 64.75\\
HKI-DDCQG(Q-7B, T5-B) & 56.69 & 49.91 & 45.07 & 41.56 & 89.81 & 48.22 & 3.78 & 55.45 & 67.84\\
HKI-DDCQG(BLIP, T5-L) & 46.48 & 39.34 & 34.54 & 30.94 & 85.18 & 39.50 & 2.80 & 45.82 & 59.25\\
HKI-DDCQG(Q-3B, T5-L) & 57.54 & 51.12 & 46.24 & 44.16 & 90.11 & 50.09 & 4.01 & 58.18 & 65.20 \\
HKI-DDCQG(Q-7B, T5-L) & 59.63 & 53.06 & 48.00 & 44.85 & 90.98 & 50.73 & 4.04 & 59.57 & 69.31 \\
\hline
\end{tabular}
\end{adjustbox}
\label{table:2}
\vspace{-12pt}
\end{table*}
\begin{table}[t!]
\small
\centering
\caption{Evaluation results on K-VQG dataset.}
\begin{adjustbox}{width=\columnwidth}
\begin{tabular}{ccccc}
\hline
Dataset & Model & BLEU-4 & CIDEr & METEOR \\
\hline
\multirow{5}{*}{K-VQG} & IM-VQG  & 11.44 & 17.07 & 0.26 \\
~ & K-VQG & 18.84 & 22.79 & 1.31 \\
~ & ConVQG & 20.01 & 22.66 & 1.53 \\
~ & $\text{HKI-DDCQG}_{B}$  & 25.34 & 26.52 & 1.82 \\
~ & $\text{HKI-DDCQG}_{Q}$ & 32.12 & 30.12 & 2.08 \\
\hline
\end{tabular}
\end{adjustbox}
\label{tab:compare_results_2}
\vspace{-15pt}
\end{table}
\subsection{Baselines}
VQG approaches can be classified into two main categories: \textit{Fine-tuned} and \textit{In-Context Learning} methods. 
\textit{Fine-tuned} methods use models like BERT \cite{devlin2018bert}, T5 \cite{raffel2020exploring}, or GPT-2 \cite{radford2019language} for question generation, with examples including IM-VQG \cite{krishna2019information}, K-VQG \cite{uehara2023k}, Patch-TRM \cite{lu2021iconqa}, ConVQG \cite{mi2024convqg}, LV2-Net \cite{liu-etal-2024-look}, and KC-VQG \cite{liu-etal-2024-knowledge}. 
\textit{In-Context Learning} methods utilize large vision-language models to extract relevant information from diagrams and generate questions using three reference questions.
This includes open-source models like Qwen2.5-VL \cite{bai2025qwen2}, MiniCPM-V \cite{yao2024minicpm}, DeepSeek-VL \cite{lu2024deepseek}, and InternVL2.5\cite{wang2025internvideo2}, as well as closed-source models such as GLM4-V pro, Claude-3.5 Sonnect, and GPT-4o.

\subsection{Evaluation metrics}
We evaluate our model using established language generation metrics, including BLEU \cite{papineni2002bleu}, Bert Score, METEOR \cite{denkowski2014meteor}, CIDEr \cite{vedantam2015cider}, and ROUGE \cite{lin2004rouge}.
We evaluate model performance using the \textit{pycocoevalcap} package to measure alignment between generated and ground truth questions.
At the same time, in order to evaluate the relevance of generated questions to diagrams, we introduce the FLEUR evaluation index \cite{lee2024fleur} to supplement the evaluation metrics.

\subsection{Results}
We comprehensively evaluate various models on our DiagramQG dataset, as shown in Table \ref{table:2}.
Among the in-context learning methods that utilize Vision-Language Models (VLMs), models with larger parameter counts generally perform significantly better, indicating their abundant knowledge across different subjects.
Interestingly, Qwen2.5-VL 3B achieves impressive and competitive results despite having fewer parameters, making it a suitable base model for further comprehensive experiments.
Tests conducted with GPT-4o and Claude-3.5-Sonnet using only text inputs result in reduced generation performance, highlighting the critical importance of incorporating diagrams in the question generation process.
Among the fine-tuned methods, KVQG and Patch-TRM do not leverage VLMs for additional knowledge, while ConVQG and KC-VQG do.
The latter models demonstrate superior performance, with KC-VQG showing improved results as VLM parameters scale up.
By observing the Fleur metrics, we find that T5-Base and T5-Large do not show significant improvement in image information understanding, indicating that VLM's contribution to the correlation between the image and the question is more obvious than T5.
HKI-DDCQG framework achieves the best performance across all quantitative evaluation metrics, regardless of whether it is based on T5-Base or T5-Large, particularly when using the larger-scale Qwen2.5-VL \cite{bai2025qwen2} model.
These impressive results convincingly validate HKI-DDCQG as an effective baseline for future research by demonstrating the value of incorporating subject knowledge in the DDCQG task.

\begin{table}[t!]
\small
\centering
\caption{Results on three VQG datasets.}
\begin{adjustbox}{width=\columnwidth}
\begin{tabular}{ccccc}
\hline
Dataset & Model & BLEU-1 & CIDEr & METEOR \\
\hline
\multirow{5}{*}{VQG-COCO} &  MDN & 36.0 & 23.4 & 0.51 \\
~ & MC-BMN & 40.7 & 22.6 & 0.50 \\
~ & ConVQG  & 50.2 & 26.4 & 0.56 \\
~ & $\text{HKI-DDCQG}_{B}$  & 55.6 & 43.4 & 0.74 \\
~ & $\text{HKI-DDCQG}_{Q}$ & 58.7 & 47.6 & 1.08 \\
\hline
\multirow{5}{*}{OK-VQA} &  IM-VQG  & 36.47 & 30.25 & 0.15\\
~ & KVQG  & 27.18 & 55.38 & 0.13\\
~ & LV2-Net  & 29.90 & 92.17 & 0.15\\
~ & $\text{HKI-DDCQG}_{B}$ & 31.51 & 112.61 & 0.20\\
~ & $\text{HKI-DDCQG}_{Q}$ & 33.82 & 128.31 & 0.24\\
\hline
\multirow{5}{*}{A-OKVQA} & IM-VQG  & 39.30 &  22.11  & 0.12\\
~ & KVQG & 30.56 & 40.97 & 0.13\\
~ & LV2-Net  & 32.11 & 60.06 & 0.14\\
~ & $\text{HKI-DDCQG}_{B}$ & 33.21 & 78.42 & 0.18\\
~ & $\text{HKI-DDCQG}_{Q}$  & 35.31 & 93.56  & 0.22\\
\hline
\end{tabular}
\end{adjustbox}
\label{tab:compare_results}
\vspace{-5pt}
\end{table}

\subsection{Results on Other Dataset}
We evaluate HKI-DDCQG's generalization capabilities on four natural image VQG datasets: VQG-COCO, OK-VQA, A-OKVQA, and K-VQG (Tables \ref{tab:compare_results_2} and \ref{tab:compare_results}). 
Our model consistently outperforms existing approaches, particularly in CIDEr and METEOR metrics, indicating superior semantic comprehension and contextual coherence.
\begin{table}[t]
\caption{Results on DiagramQG with Qwen2.5-VL 3B \& T5-Base, where H.K.E, K.S and K.F mean HierKnowExtract, KnowSelect and KnowFusionQG.}
\begin{adjustbox}{width=\columnwidth}
\centering
\small
\begin{tabular}{cccccc}
\hline
H.K.E &  K.S & K.F & BLEU-4 & METEOR & CIDEr \\
\hline
$\checkmark$ & $\checkmark$ & $\checkmark$ & 40.90 & 47.61 & 3.74 \\
$\checkmark$ & $\checkmark$ &  & 38.85 & 45.42 & 3.48\\
$\checkmark$ &  & $\checkmark$ & 35.57 & 44.02 & 2.95\\
$\checkmark$ &  &  & 34.04 & 42.52 & 2.82\\
  & & $\checkmark$ & 14.32 & 25.64 & 0.48\\
\hline
\end{tabular}
\end{adjustbox}
\vspace{-5pt}
\label{table:100}
\end{table}
\subsection{Ablation experiment}
Since \textbf{KnowSelect} inherently builds upon \textbf{HierKnowExtract}, we cannot evaluate it in isolation. 
Ablation results (Table \ref{table:100}) demonstrate that knowledge incorporation systematically enhances question generation quality. 
\textbf{KnowSelect} shows greater impact than \textbf{KnowFusionQG}, indicating its effectiveness in filtering and retaining relevant information from \textbf{HierKnowExtract}'s output.

\begin{table}[t]
\centering
\small
\caption{Results on DiagramQG under different settings. We employ HKI-DDCQG (Qwen2.5-VL 3B \& T5-Base) to obtain the results.}
\begin{tabular}{ccccc}
\hline
Setting &  BLEU-4 & METEOR & CIDEr \\
\hline
n=1,m=4 & 35.45 & 41.70 & 2.98  \\
n=2,m=4 & 38.13 & 46.01 & 3.28  \\
n=4,m=4 & 40.63 & 47.25 & 3.72  \\
\hline
n=3,m=1 & 37.14 & 44.32 & 3.35  \\
n=3,m=2 & 38.85 & 45.63 & 3.50  \\
n=3,m=4 & 41.16 & 47.74 & 3.77  \\
n=3,m=5 & 41.41 & 48.12 & 3.78  \\
\hline
n=3,m=3 & 40.90 & 47.61 & 3.74 \\
\hline
\end{tabular}
\label{table:5}
\end{table}
\subsection{Parameter Sensitive Study}
We examine two parameters: the number of pyramid layers ($n$) for diagram division and the number of knowledge sentences ($m$) selected for question generation. Results in Table \ref{table:5} show optimal performance at $n=3$, with finer divisions adding noise in knowledge extraction.
However, performance continues to improve after $m=3$, although the marginal gains come at the cost of increased input length and computational overhead.
\begin{figure}[t]
    \centering
    \includegraphics[width=0.5\textwidth]{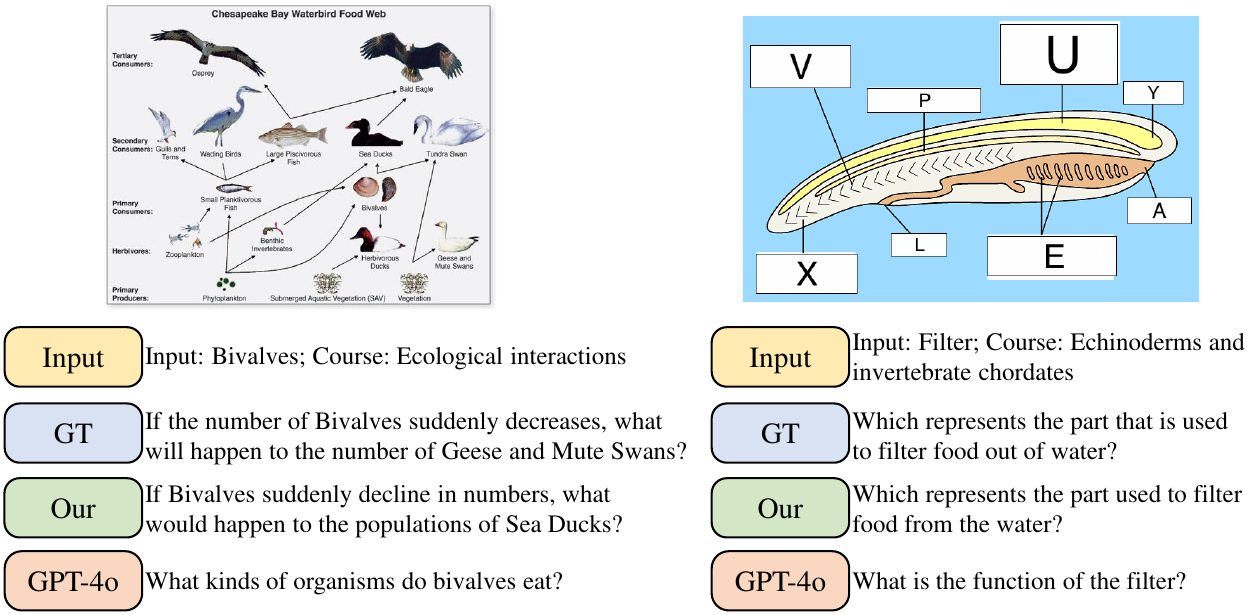}
    \vspace{-15pt}
    \caption{Comparison between HKI-DDCQG and GPT-4o using two examples from the DiagramQG dataset.}
    \label{fig:7}
    \vspace{-10pt}
\end{figure}
\subsection{Case Study}
Figure \ref{fig:7} compares HKI-DDCQG with GPT-4o on DiagramQG examples. 
While GPT-4o generates fluent questions, HKI-DDCQG demonstrates superior capability in producing course-relevant questions that better assess student understanding. 
More cases are provided in supplementary materials.
\section{Conclusion}
We present DiagramQG, the first diagram-specific question generation dataset, and HKI-DDCQG, a framework leveraging hierarchical knowledge integration for diagram-driven course question generation. 
Experimental results demonstrate HKI-DDCQG's superior performance over existing VQG and vision-language models, advancing intelligent education \cite{misra2018learning} through automated, personalized question generation.

\section{Acknowledgements}
This work was supported by the National Key Research and Development Program of China (2022YFC3303600), National Natural Science Foundation of China (No. 62137002, 62293550, 62293553, 62293554, 62450005, 62437002, 62477036, 62477037, 62176209, 62192781, 62306229),  `LENOVO-XJTU' Intelligent Industry Joint Laboratory Project, the Shaanxi Provincial Social Science Foundation Project (No. 2024P041), the Natural Science Basic Research Program of Shaanxi (No. 2023-JC-YB-593), the National Research Foundation, Singapore, under its NRF Fellowship (Award\# NRF-NRFF14-2022-0001), the Youth Innovation Team of Shaanxi Universities `Multi-modal Data Mining and Fusion', Project of China Knowledge Centre for Engineering Science and Technology, and the Youth AI Talents Fund of China Association of Automation (Grant No. HBRC-JKYZD-2024-311). 

\section{Limitation}
Our DiagramQG dataset and HKI-DDCQG framework, while advancing educational question generation, face two main limitations. First, our approach generates questions without multiple-choice options. Adding the capability to generate both questions and plausible distractors would enhance the system's educational value, as multiple-choice questions effectively assess student understanding across knowledge levels. However, DiagramQG also establishes the groundwork for future research in the simultaneous generation of diagram-based questions and distractor options.
Second, the long-tail distribution across 371 courses poses challenges for uniform performance. Models may show lower effectiveness for underrepresented courses with fewer training examples. However, this distribution reflects real-world educational patterns, and our framework maintains strong generalization across diverse domains.
Despite these constraints, DiagramQG significantly advances educational AI research, providing a robust foundation for diagram-driven course question generation. 

\section{Ethical Statement}
In developing DiagramQG, we prioritized ethical considerations throughout the entire process.
Our data collection strictly adhered to copyright guidelines, utilizing only publicly available educational resources with appropriate permissions or fair use provisions.
All data comply with CC BY-SA 4.0, CC BY-NC-SA, and MIT licenses and have been reviewed to ensure educational value and accuracy. 
We will publicly release complete datasets under appropriate licenses, both reducing unnecessary carbon footprint and optimizing processing pipelines to lower computational overhead.
We fully recognize the broad impacts of automated question-generation systems.
Our work aims to assist educators, with these systems designed to complement.
To address representation bias, we constructed a diverse dataset spanning 6 disciplines, 37 subjects, and 371 courses. 
While natural imbalances exist in educational content availability, our dataset provides broader coverage than previous work, enabling contextually appropriate question generation across various educational domains.
As an important resource driving AI capabilities in educational reasoning, DiagramQG maintains high standards for data quality and ethical considerations.
In all experiments, we strictly comply with all licensing requirements for models and data.

\bibliography{acl_latex}

\begin{thebibliography}{48}
\providecommand{\natexlab}[1]{#1}

\bibitem[{Bai et~al.(2025)Bai, Chen, Liu, Wang, Ge, Song, Dang, Wang, Wang, Tang et~al.}]{bai2025qwen2}
Shuai Bai, Keqin Chen, Xuejing Liu, Jialin Wang, Wenbin Ge, Sibo Song, Kai Dang, Peng Wang, Shijie Wang, Jun Tang, and 1 others. 2025.
\newblock Qwen2. 5-vl technical report.
\newblock \emph{arXiv preprint arXiv:2502.13923}.

\bibitem[{Bi et~al.(2022)Bi, Wang, Xue, Chen, and Huang}]{bi2022inferential}
Chao Bi, Shuhui Wang, Zhe Xue, Shengbo Chen, and Qingming Huang. 2022.
\newblock Inferential visual question generation.
\newblock In \emph{Proceedings of the 30th ACM International Conference on Multimedia}, pages 4164--4174.

\bibitem[{Chen et~al.(2023)Chen, Guo, Xie, Cai, and Li}]{chen2023deconfounded}
Jiali Chen, Zhenjun Guo, Jiayuan Xie, Yi~Cai, and Qing Li. 2023.
\newblock Deconfounded visual question generation with causal inference.
\newblock In \emph{Proceedings of the 31st ACM International Conference on Multimedia}, pages 5132--5142.

\bibitem[{Chen et~al.(2024)Chen, Qin, Zhang, Chen, Xu, and Che}]{chen2024m3cot}
Qiguang Chen, Libo Qin, Jin Zhang, Zhi Chen, Xiao Xu, and Wanxiang Che. 2024.
\newblock M3cot: A novel benchmark for multi-domain multi-step multi-modal chain-of-thought.
\newblock In \emph{Proceedings of the 62nd Annual Meeting of the Association for Computational Linguistics (Volume 1: Long Papers)}, pages 8199--8221.

\bibitem[{Cook(2006)}]{cook2006visual}
Michelle~Patrick Cook. 2006.
\newblock Visual representations in science education: The influence of prior knowledge and cognitive load theory on instructional design principles.
\newblock \emph{Science education}, 90(6):1073--1091.

\bibitem[{Denkowski and Lavie(2014)}]{denkowski2014meteor}
Michael Denkowski and Alon Lavie. 2014.
\newblock Meteor universal: Language specific translation evaluation for any target language.
\newblock In \emph{Proceedings of the ninth workshop on statistical machine translation}, pages 376--380.

\bibitem[{Devlin(2018)}]{devlin2018bert}
Jacob Devlin. 2018.
\newblock Bert: Pre-training of deep bidirectional transformers for language understanding.
\newblock \emph{arXiv preprint arXiv:1810.04805}.

\bibitem[{Fan et~al.(2018)Fan, Wei, Li, Lan, and Huang}]{fan2018question}
Zhihao Fan, Zhongyu Wei, Piji Li, Yanyan Lan, and Xuanjing Huang. 2018.
\newblock A question type driven framework to diversify visual question generation.
\newblock In \emph{IJCAI}, pages 4048--4054.

\bibitem[{G{\'o}mez-P{\'e}rez and Ortega(2020)}]{gomez2020isaaq}
Jos{\'e}~Manuel G{\'o}mez-P{\'e}rez and Ra{\'u}l Ortega. 2020.
\newblock Isaaq-mastering textbook questions with pre-trained transformers and bottom-up and top-down attention.
\newblock In \emph{Proceedings of the 2020 Conference on Empirical Methods in Natural Language Processing (EMNLP)}, pages 5469--5479.

\bibitem[{Huang et~al.(2025{\natexlab{a}})Huang, Lai, Zhang, Wu, Ma, Zhang, and Liu}]{huang2025evochart}
Muye Huang, Han Lai, Xinyu Zhang, Wenjun Wu, Jie Ma, Lingling Zhang, and Jun Liu. 2025{\natexlab{a}}.
\newblock Evochart: A benchmark and a self-training approach towards real-world chart understanding.
\newblock In \emph{Proceedings of the AAAI Conference on Artificial Intelligence}, volume~39, pages 3680--3688.

\bibitem[{Huang et~al.(2025{\natexlab{b}})Huang, Zhang, Lai, Wu, Zhang, and Liu}]{huang2025vprochart}
Muye Huang, Lingling Zhang, Han Lai, Wenjun Wu, Xinyu Zhang, and Jun Liu. 2025{\natexlab{b}}.
\newblock Vprochart: Answering chart question through visual perception alignment agent and programmatic solution reasoning.
\newblock In \emph{Proceedings of the AAAI Conference on Artificial Intelligence}, volume~39, pages 3689--3696.

\bibitem[{Kembhavi et~al.(2017)Kembhavi, Seo, Schwenk, Choi, Farhadi, and Hajishirzi}]{kembhavi2017you}
Aniruddha Kembhavi, Minjoon Seo, Dustin Schwenk, Jonghyun Choi, Ali Farhadi, and Hannaneh Hajishirzi. 2017.
\newblock Are you smarter than a sixth grader? textbook question answering for multimodal machine comprehension.
\newblock In \emph{Proceedings of the IEEE Conference on Computer Vision and Pattern recognition}, pages 4999--5007.

\bibitem[{Krishna et~al.(2019)Krishna, Bernstein, and Fei-Fei}]{krishna2019information}
Ranjay Krishna, Michael Bernstein, and Li~Fei-Fei. 2019.
\newblock Information maximizing visual question generation.
\newblock In \emph{Proceedings of the IEEE/CVF conference on computer vision and pattern recognition}, pages 2008--2018.

\bibitem[{Lambertus et~al.(2008)}]{lambertus2008students}
Amanda~Jane Lambertus and 1 others. 2008.
\newblock Students' understanding of the function concept: Concept images and concept definitions.

\bibitem[{Lee et~al.(2024)Lee, Park, and Kang}]{lee2024fleur}
Yebin Lee, Imseong Park, and Myungjoo Kang. 2024.
\newblock Fleur: An explainable reference-free evaluation metric for image captioning using a large multimodal model.
\newblock In \emph{Proceedings of the 62nd Annual Meeting of the Association for Computational Linguistics (Volume 1: Long Papers)}, pages 3732--3746.

\bibitem[{Li et~al.(2022)Li, Li, Xiong, and Hoi}]{li2022blip}
Junnan Li, Dongxu Li, Caiming Xiong, and Steven Hoi. 2022.
\newblock Blip: Bootstrapping language-image pre-training for unified vision-language understanding and generation.
\newblock In \emph{International conference on machine learning}, pages 12888--12900. PMLR.

\bibitem[{Lin(2004)}]{lin2004rouge}
Chin-Yew Lin. 2004.
\newblock Rouge: A package for automatic evaluation of summaries.
\newblock In \emph{Text summarization branches out}, pages 74--81.

\bibitem[{Liu et~al.(2024{\natexlab{a}})Liu, Wang, Xie, Chen, Fang, and Cai}]{liu-etal-2024-knowledge}
Hongfei Liu, Guohua Wang, Jiayuan Xie, Jiali Chen, Wenhao Fang, and Yi~Cai. 2024{\natexlab{a}}.
\newblock \href {https://aclanthology.org/2024.lrec-main.861} {Knowledge-guided cross-topic visual question generation}.
\newblock In \emph{Proceedings of the 2024 Joint International Conference on Computational Linguistics, Language Resources and Evaluation (LREC-COLING 2024)}, pages 9854--9864, Torino, Italia. ELRA and ICCL.

\bibitem[{Liu et~al.(2024{\natexlab{b}})Liu, Wang, Xie, Chen, Fang, and Cai}]{liu2024knowledge}
Hongfei Liu, Guohua Wang, Jiayuan Xie, Jiali Chen, Wenhao Fang, and Yi~Cai. 2024{\natexlab{b}}.
\newblock Knowledge-guided cross-topic visual question generation.
\newblock In \emph{Proceedings of the 2024 Joint International Conference on Computational Linguistics, Language Resources and Evaluation (LREC-COLING 2024)}, pages 9854--9864.

\bibitem[{Liu et~al.(2024{\natexlab{c}})Liu, Guo, Zhang, Liu, Zhao, Yu, and Yuan}]{liu-etal-2024-look}
Xumeng Liu, Wenya Guo, Ying Zhang, Xubo Liu, Yu~Zhao, Shenglong Yu, and Xiaojie Yuan. 2024{\natexlab{c}}.
\newblock \href {https://aclanthology.org/2024.lrec-main.943} {Look before you leap: Dual logical verification for knowledge-based visual question generation}.
\newblock In \emph{Proceedings of the 2024 Joint International Conference on Computational Linguistics, Language Resources and Evaluation (LREC-COLING 2024)}, pages 10802--10812, Torino, Italia. ELRA and ICCL.

\bibitem[{Lu et~al.(2024)Lu, Liu, Zhang, Wang, Dong, Liu, Sun, Ren, Li, Sun et~al.}]{lu2024deepseek}
Haoyu Lu, Wen Liu, Bo~Zhang, Bingxuan Wang, Kai Dong, Bo~Liu, Jingxiang Sun, Tongzheng Ren, Zhuoshu Li, Yaofeng Sun, and 1 others. 2024.
\newblock Deepseek-vl: towards real-world vision-language understanding.
\newblock \emph{arXiv preprint arXiv:2403.05525}.

\bibitem[{Lu et~al.(2022)Lu, Mishra, Xia, Qiu, Chang, Zhu, Tafjord, Clark, and Kalyan}]{lu2022learn}
Pan Lu, Swaroop Mishra, Tanglin Xia, Liang Qiu, Kai-Wei Chang, Song-Chun Zhu, Oyvind Tafjord, Peter Clark, and Ashwin Kalyan. 2022.
\newblock Learn to explain: Multimodal reasoning via thought chains for science question answering.
\newblock \emph{Advances in Neural Information Processing Systems}, 35:2507--2521.

\bibitem[{Lu et~al.(2021)Lu, Qiu, Chen, Xia, Zhao, Zhang, Yu, Liang, and Zhu}]{lu2021iconqa}
Pan Lu, Liang Qiu, Jiaqi Chen, Tony Xia, Yizhou Zhao, Wei Zhang, Zhou Yu, Xiaodan Liang, and Song-Chun Zhu. 2021.
\newblock Iconqa: A new benchmark for abstract diagram understanding and visual language reasoning.
\newblock \emph{arXiv preprint arXiv:2110.13214}.

\bibitem[{Luo et~al.(2024)Luo, Deng, Shen, Ng, and Chua}]{luo-etal-2024-chain}
Haohao Luo, Yang Deng, Ying Shen, See-Kiong Ng, and Tat-Seng Chua. 2024.
\newblock \href {https://doi.org/10.18653/v1/2024.acl-long.432} {Chain-of-exemplar: Enhancing distractor generation for multimodal educational question generation}.
\newblock In \emph{Proceedings of the 62nd Annual Meeting of the Association for Computational Linguistics (Volume 1: Long Papers)}, pages 7978--7993, Bangkok, Thailand. Association for Computational Linguistics.

\bibitem[{Ma et~al.(2022)Ma, Chai, Huang, Liu, You, and Zheng}]{ma2022weakly}
Jie Ma, Qi~Chai, Jingyue Huang, Jun Liu, Yang You, and Qinghua Zheng. 2022.
\newblock Weakly supervised learning for textbook question answering.
\newblock \emph{IEEE Transactions on Image Processing}, 31:7378--7388.

\bibitem[{Mi et~al.(2024)Mi, Montariol, Navarro, Dai, Bosselut, and Tuia}]{mi2024convqg}
Li~Mi, Syrielle Montariol, Javiera~Castillo Navarro, Xianjie Dai, Antoine Bosselut, and Devis Tuia. 2024.
\newblock Convqg: Contrastive visual question generation with multimodal guidance.
\newblock In \emph{Proceedings of the AAAI Conference on Artificial Intelligence}, volume~38, pages 4207--4215.

\bibitem[{Misra et~al.(2018)Misra, Girshick, Fergus, Hebert, Gupta, and Van Der~Maaten}]{misra2018learning}
Ishan Misra, Ross Girshick, Rob Fergus, Martial Hebert, Abhinav Gupta, and Laurens Van Der~Maaten. 2018.
\newblock Learning by asking questions.
\newblock In \emph{Proceedings of the IEEE Conference on Computer Vision and Pattern Recognition}, pages 11--20.

\bibitem[{Papineni et~al.(2002)Papineni, Roukos, Ward, and Zhu}]{papineni2002bleu}
Kishore Papineni, Salim Roukos, Todd Ward, and Wei-Jing Zhu. 2002.
\newblock Bleu: a method for automatic evaluation of machine translation.
\newblock In \emph{Proceedings of the 40th annual meeting of the Association for Computational Linguistics}, pages 311--318.

\bibitem[{Radford et~al.(2021)Radford, Kim, Hallacy, Ramesh, Goh, Agarwal, Sastry, Askell, Mishkin, Clark et~al.}]{radford2021learning}
Alec Radford, Jong~Wook Kim, Chris Hallacy, Aditya Ramesh, Gabriel Goh, Sandhini Agarwal, Girish Sastry, Amanda Askell, Pamela Mishkin, Jack Clark, and 1 others. 2021.
\newblock Learning transferable visual models from natural language supervision.
\newblock In \emph{International conference on machine learning}, pages 8748--8763. PMLR.

\bibitem[{Radford et~al.(2019)Radford, Wu, Child, Luan, Amodei, Sutskever et~al.}]{radford2019language}
Alec Radford, Jeffrey Wu, Rewon Child, David Luan, Dario Amodei, Ilya Sutskever, and 1 others. 2019.
\newblock Language models are unsupervised multitask learners.
\newblock \emph{OpenAI blog}, 1(8):9.

\bibitem[{Raffel et~al.(2020)Raffel, Shazeer, Roberts, Lee, Narang, Matena, Zhou, Li, and Liu}]{raffel2020exploring}
Colin Raffel, Noam Shazeer, Adam Roberts, Katherine Lee, Sharan Narang, Michael Matena, Yanqi Zhou, Wei Li, and Peter~J Liu. 2020.
\newblock Exploring the limits of transfer learning with a unified text-to-text transformer.
\newblock \emph{Journal of machine learning research}, 21(140):1--67.

\bibitem[{Uehara and Harada(2023)}]{uehara2023k}
Kohei Uehara and Tatsuya Harada. 2023.
\newblock K-vqg: Knowledge-aware visual question generation for common-sense acquisition.
\newblock In \emph{Proceedings of the IEEE/CVF Winter Conference on Applications of Computer Vision}, pages 4401--4409.

\bibitem[{Uehara et~al.(2018)Uehara, Tejero-De-Pablos, Ushiku, and Harada}]{uehara2018visual}
Kohei Uehara, Antonio Tejero-De-Pablos, Yoshitaka Ushiku, and Tatsuya Harada. 2018.
\newblock Visual question generation for class acquisition of unknown objects.
\newblock In \emph{Proceedings of the European conference on computer vision (ECCV)}, pages 481--496.

\bibitem[{Vedantam et~al.(2015)Vedantam, Lawrence~Zitnick, and Parikh}]{vedantam2015cider}
Ramakrishna Vedantam, C~Lawrence~Zitnick, and Devi Parikh. 2015.
\newblock Cider: Consensus-based image description evaluation.
\newblock In \emph{Proceedings of the IEEE conference on computer vision and pattern recognition}, pages 4566--4575.

\bibitem[{Wang et~al.(2022)Wang, Zhang, Luo, Yang, Hu, Qin, and Liu}]{wang2022computer}
Shaowei Wang, Lingling Zhang, Xuan Luo, Yi~Yang, Xin Hu, Tao Qin, and Jun Liu. 2022.
\newblock Computer science diagram understanding with topology parsing.
\newblock \emph{ACM Transactions on Knowledge Discovery from Data (TKDD)}, 16(6):1--20.

\bibitem[{Wang et~al.(2025{\natexlab{a}})Wang, Zhang, Wu, Qin, Zhang, and Liu}]{10812784}
Shaowei Wang, Lingling Zhang, Wenjun Wu, Tao Qin, Xinyu Zhang, and Jun Liu. 2025{\natexlab{a}}.
\newblock \href {https://doi.org/10.1109/TMM.2024.3521744} {Alignment-guided self-supervised learning for diagram question answering}.
\newblock \emph{IEEE Transactions on Multimedia}, 27:2141--2154.

\bibitem[{Wang et~al.(2024)Wang, Zhang, Zhu, Qin, Yap, Zhang, and Liu}]{wang2024cvpr}
Shaowei Wang, Lingling Zhang, Longji Zhu, Tao Qin, Kim-Hui Yap, Xinyu Zhang, and Jun Liu. 2024.
\newblock Cog-dqa: Chain-of-guiding learning with large language models for diagram question answering.
\newblock In \emph{Proceedings of the IEEE Conference on Computer Vision and Pattern Recognition (CVPR)}.

\bibitem[{Wang et~al.(2025{\natexlab{b}})Wang, Li, Yan, He, Yu, Zeng, Wang, Ma, Huang, Gao et~al.}]{wang2025internvideo2}
Yi~Wang, Xinhao Li, Ziang Yan, Yinan He, Jiashuo Yu, Xiangyu Zeng, Chenting Wang, Changlian Ma, Haian Huang, Jianfei Gao, and 1 others. 2025{\natexlab{b}}.
\newblock Internvideo2. 5: Empowering video mllms with long and rich context modeling.
\newblock \emph{arXiv preprint arXiv:2501.12386}.

\bibitem[{Xie et~al.(2025)Xie, Cheng, Zhang, Cai, Hu, Xie, and Li}]{xie2025explicitly}
Jiayuan Xie, Mengqiu Cheng, Xinting Zhang, Yi~Cai, Guimin Hu, Mengying Xie, and Qing Li. 2025.
\newblock Explicitly guided difficulty-controllable visual question generation.
\newblock In \emph{Proceedings of the AAAI Conference on Artificial Intelligence}, volume~39, pages 25552--25560.

\bibitem[{Xie et~al.(2022)Xie, Fang, Cai, Huang, and Li}]{xie2022knowledge}
Jiayuan Xie, Wenhao Fang, Yi~Cai, Qingbao Huang, and Qing Li. 2022.
\newblock Knowledge-based visual question generation.
\newblock \emph{IEEE Transactions on Circuits and Systems for Video Technology}, 32(11):7547--7558.

\bibitem[{Xu et~al.(2023)Xu, Lin, Liu, Zhang, Zhao, Chai, Pan, Huang, and Wang}]{XU2023109588}
Fangzhi Xu, Qika Lin, Jun Liu, Lingling Zhang, Tianzhe Zhao, Qi~Chai, Yudai Pan, Yi~Huang, and Qianying Wang. 2023.
\newblock \href {https://doi.org/10.1016/j.patcog.2023.109588} {Moca: Incorporating domain pretraining and cross attention for textbook question answering}.
\newblock \emph{Pattern Recognition}, 140:109588.

\bibitem[{Xu et~al.(2020)Xu, Wang, Yang, Hanjalic, and Shen}]{xu2020radial}
Xing Xu, Tan Wang, Yang Yang, Alan Hanjalic, and Heng~Tao Shen. 2020.
\newblock Radial graph convolutional network for visual question generation.
\newblock \emph{IEEE transactions on neural networks and learning systems}, 32(4):1654--1667.

\bibitem[{Yao et~al.(2024{\natexlab{a}})Yao, Yu, Zhao, Shafran, Griffiths, Cao, and Narasimhan}]{yao2024tree}
Shunyu Yao, Dian Yu, Jeffrey Zhao, Izhak Shafran, Tom Griffiths, Yuan Cao, and Karthik Narasimhan. 2024{\natexlab{a}}.
\newblock Tree of thoughts: Deliberate problem solving with large language models.
\newblock \emph{Advances in Neural Information Processing Systems}, 36.

\bibitem[{Yao et~al.(2024{\natexlab{b}})Yao, Yu, Zhang, Wang, Cui, Zhu, Cai, Li, Zhao, He et~al.}]{yao2024minicpm}
Yuan Yao, Tianyu Yu, Ao~Zhang, Chongyi Wang, Junbo Cui, Hongji Zhu, Tianchi Cai, Haoyu Li, Weilin Zhao, Zhihui He, and 1 others. 2024{\natexlab{b}}.
\newblock Minicpm-v: A gpt-4v level mllm on your phone.
\newblock \emph{arXiv preprint arXiv:2408.01800}.

\bibitem[{Yue et~al.(2024)Yue, Ni, Zhang, Zheng, Liu, Zhang, Stevens, Jiang, Ren, Sun et~al.}]{yue2024mmmu}
Xiang Yue, Yuansheng Ni, Kai Zhang, Tianyu Zheng, Ruoqi Liu, Ge~Zhang, Samuel Stevens, Dongfu Jiang, Weiming Ren, Yuxuan Sun, and 1 others. 2024.
\newblock Mmmu: A massive multi-discipline multimodal understanding and reasoning benchmark for expert agi.
\newblock In \emph{Proceedings of the IEEE/CVF Conference on Computer Vision and Pattern Recognition}, pages 9556--9567.

\bibitem[{Zhang et~al.(2025)Zhang, Dong, Wu, Huang, Jia, Fernando, Shou, Zhang, and Liu}]{zhang-etal-2025-physreason}
Xinyu Zhang, Yuxuan Dong, Yanrui Wu, Jiaxing Huang, Chengyou Jia, Basura Fernando, Mike~Zheng Shou, Lingling Zhang, and Jun Liu. 2025.
\newblock \href {https://doi.org/10.18653/v1/2025.acl-long.811} {{P}hys{R}eason: A comprehensive benchmark towards physics-based reasoning}.
\newblock In \emph{Proceedings of the 63rd Annual Meeting of the Association for Computational Linguistics (Volume 1: Long Papers)}, pages 16593--16615, Vienna, Austria. Association for Computational Linguistics.

\bibitem[{Zhang et~al.(2024{\natexlab{a}})Zhang, Zhang, Hu, Liu, Wang, and Wang}]{zhang2024alignment}
Xinyu Zhang, Lingling Zhang, Xin Hu, Jun Liu, Shaowei Wang, and Qianying Wang. 2024{\natexlab{a}}.
\newblock Alignment relation is what you need for diagram parsing.
\newblock \emph{IEEE Transactions on Image Processing}.

\bibitem[{Zhang et~al.(2024{\natexlab{b}})Zhang, Zhang, Li, Karypis, Smola et~al.}]{zhang2023multimodal}
Zhuosheng Zhang, Aston Zhang, Mu~Li, George Karypis, Alex Smola, and 1 others. 2024{\natexlab{b}}.
\newblock Multimodal chain-of-thought reasoning in language models.
\newblock \emph{Transactions on Machine Learning Research}.

\end{thebibliography}
\newpage

\section{Appendix}
In this supplementary material, we provide more details on
 our implementation and experiments as follows:
\begin{itemize}
    \item Section A: More details on implementation;
    \item Section B: More details on baselines;
    \item Section C: More details on DiagramQG;
    \item Section D: More case studies on Diagram.
\end{itemize}
\par
\subsection*{A. More details on implementation}
In our experimental setup, we developed a framework based on the T5 architecture, utilizing both its Base and Large variants respectively, alongside CLIP (ViT-B/32) for visual understanding capabilities. 
The framework's trainable parameters, excluding pre-trained components, were initialized following a normal distribution ($\mu=0$, $\sigma=0.02$). 
For optimization, we implemented the AdamW optimizer with a dual learning rate strategy: a conservative $1e-5$ for CLIP components to maintain visual feature integrity, and a higher $5e-5$ for remaining components. 
The training process spanned $20$ epochs, incorporating linear learning rate warmup during the initial $2$ epochs followed by cosine decay. 
Our implementation featured specific configurations including $256$ tokens for maximum input sequence length, $32$ tokens for output sequence length, a batch size of $32$, $4$ gradient accumulation steps, $0.01$ weight decay, and $0.1$ dropout rate, with FP16 mixed precision training enabled.
All experimental procedures were executed on a hardware setup consisting of two NVIDIA A100 80G GPUs, utilizing CUDA 11.8 and PyTorch 1.13.1.
\subsection*{B. More details on baselines}
Visual Question Generation (VQG) approaches can be classified into two main categories: fine-tuning-based methods and large vision-language models.
\par
For \textbf{Fine-tuning-based methods}, using the same data split approach as our method, the dataset is divided into training, validation, and test sets in a ratio of 70:5:25. Models are trained on the training set, validated on the validation set, and final results are reported on the test set.
\par
K-VQG \cite{uehara2023k} integrates UNITER, a multi-modal transformer, to encode both visual features (extracted via Faster R-CNN) and masked knowledge triplets. 
The model processes visual region features with positional information and combines them with tokenized knowledge triplets through a BART-based decoder, generating questions 
\par
Patch-TRM \cite{lu2021iconqa} employs a hierarchical patch-based transformer that processes diagrams by decomposing them into meaningful patches through a pyramid layout. 
The model combines ResNet and vision Transformer for visual processing, using attention mechanisms to fuse visual and textual features, enabling effective capture of both local details and global relationships for diagram-focused question generation.
\par
ConVQG \cite{mi2024convqg} introduces a contrastive learning framework with dual modality-specific objectives for visual question generation. 
By contrasting image-specific and text-guided features, the model generates diverse questions that are both image-specific and controllable through flexible textual constraints such as answers, captions, or knowledge triplets, enabling precise control over question content while maintaining visual grounding.
\par
KC-VQG \cite{liu-etal-2024-knowledge} presents a knowledge-guided framework that combines topic-aware visual attention and Large Language Model (LLM) generated knowledge for question generation. 
The model integrates three components: a topic-aware visual feature extractor, a knowledge extractor with discriminative filtering, and a GPT-2 based decoder, enabling it to generate questions that incorporate both explicit visual information and implicit commonsense knowledge about specified topics.
\par
LV2-Net \cite{liu-etal-2024-look} integrates logical verification into both knowledge acquisition and question generation processes, hence its designation as LVˆ2-Net. By performing a dual logical structure check—examining the relationships between visual content (V), attributes (A), knowledge (K), ground-truth answers, and the generated questions (Q) at two distinct stages within the knowledge-based visual question generation (KB-VQG) pipeline—LV2-Net is capable of producing a diverse range of insightful knowledge-driven visual questions.
\par
For \textbf{large vision-language models} (VLMs), whether open-source or closed-source, we selected three reference questions for each course. 
These reference questions, along with target and course textual constraints, are incorporated into the prompt during the VLM's question generation process to obtain final outputs.
\par
\begin{figure*}[t]
    \centering
    \includegraphics[width=0.98\textwidth]{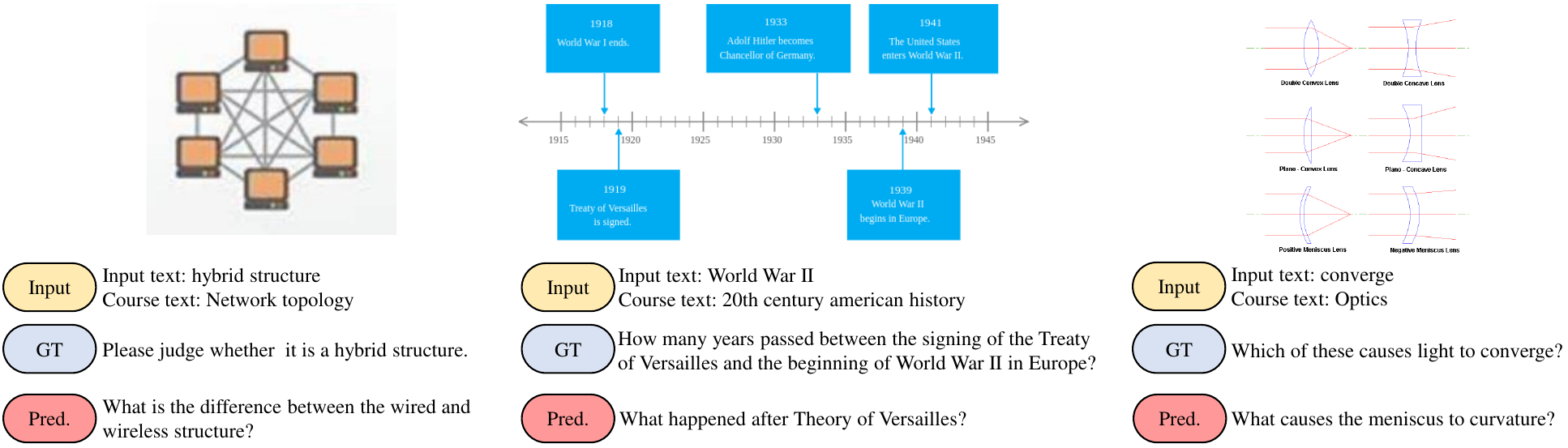}
    \caption{Three different cases ({Text Constraint Lost}) in DiagramQG dataset.}
    \label{fig:13}
\end{figure*}
\begin{figure*}[t]
    \centering
    \includegraphics[width=0.98\textwidth]{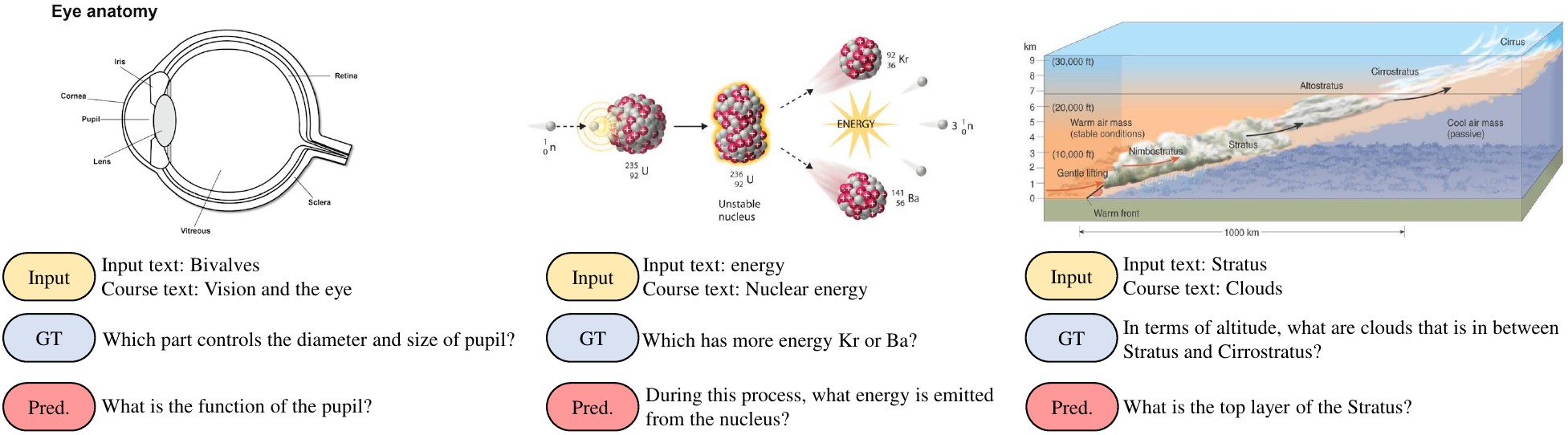}
    \caption{Three different cases ({Diagram Interpretation Bias}) in DiagramQG dataset.}
    \label{fig:14}
\end{figure*}
\begin{figure*}[t]
    \centering
    \includegraphics[width=0.98\textwidth]{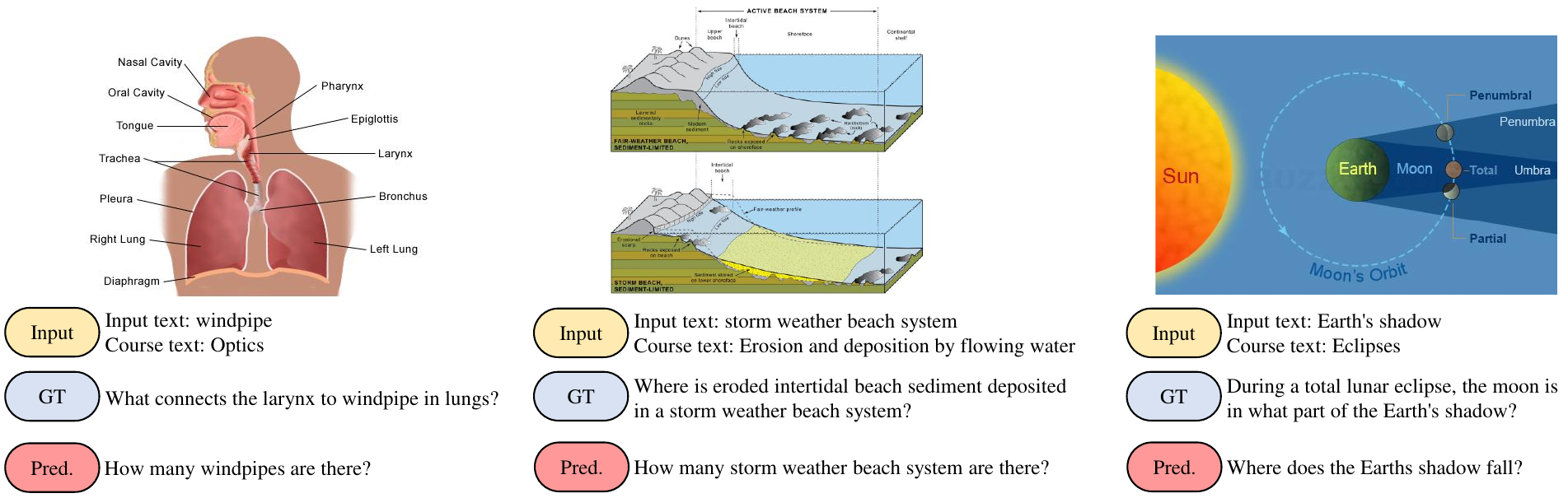}
    \caption{Three different cases ({Course Ambiguity}) in DiagramQG dataset.}
    \label{fig:15}
\end{figure*}
\textbf{Open-Source Large vision-language models:}
\par
Qwen2.5VL \cite{bai2025qwen2} is a multimodal model with key enhancements in document parsing, object grounding, and video understanding. It boasts powerful omnidocument parsing capabilities, excelling in diverse document types and languages, including complex elements like tables, charts, and formulas. The model offers precise object grounding with support for various coordinate formats, enabling advanced spatial reasoning. Its ultra-long video understanding is enhanced by dynamic resolution in the temporal dimension, allowing for comprehension of hours-long videos and fine-grained event localization. Furthermore, Qwen2.5VL features improved agent functionality for computer and mobile devices through enhanced grounding, reasoning, and decision-making. Architecturally, it incorporates dynamic FPS sampling and temporal mRoPE for video understanding, along with a streamlined vision encoder utilizing window attention, SwiGLU, and RMSNorm. Available in 3B, 7B, 32B, and 72B parameter versions, its streaming architecture ensures efficient processing of various inputs and strong performance on diagram-related tasks.
\par
MiniCPM-V \cite{yao2024minicpm} presents a compact yet effective vision-language model that combines SigLip-400M visual encoder with Qwen2-7B language model, totaling 8B parameters. 
The model offers efficient multi-image and video understanding capabilities while maintaining competitive performance on vision-language tasks through its streamlined architecture and parameter-efficient design.
\par
\begin{table}[t]
\centering
\begin{tabular}{cc}
\toprule
Statistic & Number \\
\midrule
Total  Diagram & 15,720 \\
Total  Question & 25,798 \\
Total Discipline & 6 \\
Total Subject & 37 \\
Total Course & 371 \\
\midrule
Train Diagram & 11,817 \\
Train Question & 17,880 \\
TotTrainal Discipline & 6 \\
Train Subject & 37 \\
Train Course & 351 \\
\midrule
Val Diagram & 1,151 \\
Val Question & 1,104 \\
Val Discipline & 6 \\
Val Subject & 33 \\
Val Course & 310 \\
\midrule
Test Diagram & 6,767 \\
Test Question & 5,565 \\
Test Discipline & 6 \\
Test Subject & 37 \\
Test Course & 371 \\
\bottomrule
\end{tabular}
\caption{Main statistics in DiagramQG}
\label{tab:11}
\end{table}
\par
DeepSeek-VL \cite{lu2024deepseek} introduces a balanced approach to vision-language modeling that builds upon the DeepSeek language model series. 
The model emphasizes maintaining strong language capabilities while developing visual understanding, featuring high-resolution processing capabilities and a carefully curated training strategy. 
Through its systematic scaling methodology from 1B to 7B parameters, it achieves competitive performance in practical applications while maintaining efficiency in multi-modal processing.
\par
InternVL 2.5 \cite{wang2025internvideo2} is an advanced open-source Multimodal Large Language Model (MLLM) series, building on InternVL 2.0 with enhanced training, testing, and data quality. It rivals top commercial models like GPT-4o and Claude-3.5-Sonnet, notably being the first open-source MLLM to exceed 70\% on the MMMU benchmark. Key innovations include a Progressive Scaling Strategy for efficient vision encoder and LLM alignment, improved training with Random JPEG Compression and Loss Reweighting, and Well-structured Data Organization for higher quality and training efficiency. InternVL 2.5 aims to advance the open-source multimodal AI community.
\par
\textbf{Closed-Source Large vision-language models:}
\par
GLM4-V is a large-scale multimodal language model developed by ZhiPu, featuring outstanding visual understanding and language generation capabilities. It employs a unified pre-training framework, capable of handling multiple modalities such as text, images, and audio. GLM4-V demonstrates strong performance in tasks like visual question answering, image description, and cross-modal reasoning, and can quickly adapt to new scenarios through few-shot learning. The model supports both Chinese and English, excelling in multilingual understanding and generation.
\par
Claude-3.5-Sonnet is a next-generation multimodal assistant developed by Anthropic, exhibiting exceptional performance in both visual and language understanding. It utilizes an innovative neural network architecture that allows for deep comprehension of image content and complex reasoning. The model has strict controls in terms of safety and ethics, capable of identifying and filtering inappropriate content. A notable feature is its strong contextual understanding and coherent conversational abilities.
\par
GPT-4o is the latest large language model developed by OpenAI, possessing powerful multimodal understanding and generation capabilities. It can process image and text inputs and perform complex reasoning and knowledge integration. 
This model showcases remarkable few-shot learning abilities, quickly mastering new tasks with only a few examples. GPT-4o also excels in creativity, generating high-quality text, code, and other creative content.

\subsection*{C. More details on DiagramQG}
The DiagramQG dataset is a comprehensive collection of questions covering 6 disciplines, 37 subjects, and 371 courses, consisting of 25,798 questions and 15,720 diagrams.
This dataset aims to encourage models to generate questions that assess students' course understanding by leveraging the provided input \& course text constraints and diagrams, as shown in Figure \ref{fig:1}. 
Accomplishing this task requires two key capabilities from computational models. 
First, they must demonstrate sophisticated visual comprehension abilities. 
Second, they need comprehensive domain knowledge across multiple academic disciplines.
Following established methodological practices in machine learning, we implemented a stratified partitioning of the dataset, allocating 70\% for training, 5\% for validation, and 25\% for testing purposes.
The comprehensive statistical distribution across these partitions, along with the aggregate dataset metrics, is presented in Table \ref{tab:11}.
It can be seen that some images have intersections, but this does not affect the experimental results, because its course and question, and can effectively detect overfitting, because overfitting will cause the generated questions to be inconsistent with the current requirements

\subsection*{D More case studies on Diagram}
Through extensive examination of the DDCQG (Diagram-based Question Generation) task, we have identified and analyzed three critical bottlenecks: \textbf{Text Constraint Lost}, \textbf{Diagram Interpretation Bias}, and \textbf{Course Ambiguity}. 
Detailed case studies illustrating these challenges are presented in Figures \ref{fig:13}, \ref{fig:14}, and \ref{fig:15}, respectively.
\par
The first challenge, Text Constraint Lost (Figure \ref{fig:13}), manifests when the generated questions fail to maintain fidelity to the target text, resulting in contextually inconsistent question generation. Our analysis suggests that this phenomenon stems from limitations in the model's instruction processing capabilities. 
Specifically, the neural architecture appears to inadequately preserve and integrate the target text constraints during the generation pipeline, leading to divergent outputs that, while potentially coherent, deviate from the intended textual context.
\par
The second challenge, Diagram Interpretation Bias (Figure \ref{fig:14}), reveals a notable gap between the model-generated questions and ground truth questions in terms of course application depth. 
While the generated questions demonstrate basic assessment of students' course comprehension, they often exhibit simplified or superficial understanding of course concepts. 
To investigate this phenomenon, we conducted a visualization analysis of the knowledge retrieval process during question generation, as illustrated in Figure \ref{fig:14}.
Our findings reveal a significant limitation: the retrieved background knowledge corpus lacks crucial content related to key concepts such as ``filter blood,'' thereby constraining the model's ability to generate sophisticated, course-relevant questions.
\par
The third challenge, Course Ambiguity (Figure \ref{fig:15}), represents a important limitation in current DDCQG systems. In these cases, the generated questions demonstrate minimal engagement with course-specific content, instead defaulting to surface-level expansions of the target text. This suggests a deeper architectural limitation in connecting diagram elements with relevant course concepts and pedagogical objectives.


\section{Details of human annotators}
For data annotation and evaluation, we engaged six graduate students (including both PhD and Master's students) from engineering disciplines who are also co-authors of this paper. 
All annotators possessed strong backgrounds in different subjects, making them well-qualified for this task. 
Since the annotators were co-authors actively involved in the research, no formal recruitment process or compensation was required, and they were fully aware of how the data would be used in the study. 
The annotation process focused solely on content evaluation and did not involve collecting any personal identifying information or expose annotators to any risks.
As this research involved co-authors analyzing academic content rather than external human subjects, it was determined to be exempt from formal ethics review board approval. 
The annotation work was conducted as part of regular academic research activities within our institution.  
No protected or sensitive demographic information was collected or used in this research.

\section{Details of Ai Assistants In Research Or Writing}
We used Claude-3.5-Sonnet, o1, o3-mini-high, and Deepseek-R1 to help us write code and polish the paper.
\par

\end{document}